\documentclass[journal,onecolumn]{IEEEtran}
\usepackage[utf8]{inputenc}
\usepackage{amsmath}
\usepackage{amssymb}
\ifCLASSINFOpdf \else
  \usepackage[dvips]{graphicx}
\fi
\usepackage{url}

\usepackage{graphicx}

\usepackage{color}
\usepackage{txfonts} 

\usepackage{textcomp} 
\usepackage{multirow}
\usepackage{makecell} 

\makeatletter
\renewcommand{\maketag@@@}[1]{\hbox{\m@th\normalsize\normalfont#1}} 
\makeatother

\newtheorem{assumption}{Assumption}
\begin{document}

\title{Robust Adaptive Filtering Based on Exponential Functional Link Network}

\author{Tao Yu,
        Wenqi Li,
        Yi Yu,~\IEEEmembership{Member,~IEEE,}
        and Rodrigo C. de Lamare~\IEEEmembership{Senior Member,~IEEE}
\thanks{Manuscript received October 27, 2020. This work was supported in part by the NNSFC under Grants 61901400 and 61433011, and in part by the Scientific Research Starting Project of SWPU under Grant 2019QHZ015. {\it (Corresponding author: Tao Yu.)}}
\thanks{Tao Yu and Wenqi Li are with the School of Electrical Engineering and Information, Southwest Petroleum University, Chengdu 610500, China (e-mail: yutao@swpu.edu.cn; wqli\_edu@163.com).}
\thanks{Yi Yu is with the School of Information Engineering, Robot Technology Used for Special Environment Key Laboratory of Sichuan Province, Southwest University of Science and Technology, Mianyang 621010, China (e-mail: yuyi\_xyuan@163.com).}
\thanks{Rodrigo C. de Lamare is with the Centre for Telecommunications Studies, Pontifical Catholic University of Rio de Janeiro, Rio de Janeiro 22451-900, Brazil, and also with the Department of Electronic Engineering, University of York, York YO10 5DD, UK (e-mail: delamare@cetuc.puc-rio.br).}
\vspace{-15pt}
}

\markboth{IEEE Transactions on Circuits and Systems II: Express Briefs}
{YU \MakeLowercase{\textit{et al.}}: Robust Adaptive Filtering Based on Exponential Functional Link Network}

\maketitle

\begin{abstract}
The exponential functional link network (EFLN) has been recently investigated and applied to nonlinear filtering. This brief proposes an adaptive EFLN filtering algorithm based on a novel inverse square root (ISR) cost function, called the EFLN-ISR algorithm, whose learning capability is robust under impulsive interference. The steady-state performance of EFLN-ISR is rigorously derived and then confirmed by numerical simulations. Moreover, the validity of the proposed EFLN-ISR algorithm is justified by the actually experimental results with the application to hysteretic nonlinear system identification.
\end{abstract}


\section{Introduction}

\IEEEPARstart{T}{he} linear-in-parameters nonlinear filtering
techniques, which possess the low computational complexity and
efficient learning capability, have been applied to diverse areas
such as the nonlinear system identification, nonlinear acoustic echo
cancellation (AEC), and nonlinear active noise control
\cite{Comminiello2018Adaptive}, and shown advantages over linear
approaches
\cite{jidf,jidf_echo,ccg,jio,intadap,saalt,jiols,ccmjio,wlmwf,wljio,jiomimo,sjidf,saabf,rrser,jiostap,locsme,oskpme,dce,mskaesprit,damdc,rdrls,dlmm}.
The most traditional nonlinear filters among them are the
trigonometric functional link network-based filter (TFLN)
\cite{Sicuranza2012BIBO}, and the second-order Volterra filter
(SOVF) \cite{Tan2001Adaptive}. Such input signals are expanded by
the trigonometric basis functions and the Volterra series to achieve
nonlinear mapping, respectively.

The conventional TFLN consisting of pure trigonometric polynomial
functions can effectively model the typical nonlinearities. It has
also been proved that the trigonometric functions with exponentially
varying amplitudes can further enhance the nonlinear modeling
capability \cite{Jensen2004Perceptual,Hermus2005Perceptual}.
Recently, a class of exponential functional link network
(EFLN)-based nonlinear filters has been constructed in
\cite{Patel2016Design}, which not only updates the weights but also
updates an exponential factor controlling the decay rate of
trigonometric basis functions. The least mean-square (LMS) approach
has been utilized to adapt the weights and the exponential factor of
EFLN-based filter, called the EFLN-LMS algorithm, where its
convergence behavior and performance analysis have been discussed in
\cite{Patel2020Convergence}. In the feedback cancellation scenarios,
the works in \cite{Vasundhara2018Decorrelated} and
\cite{Le2018Generalized} provided the EFLN-based filtering
structures to enhance the nonlinear noise mitigation capability.
Later, to improve the convergence performance and the modeling
accuracy, some improved versions of EFLN have been developed and
shown to provide better performance
\cite{Deb2020Design,Bhattacharjee2020Nonlinear}.

However, it is noteworthy that the impulsive interference may often
lead to severe performance degradation
\cite{Yu2020DCD,Yu2021Robust}. In order to perform well in the case
of impulsive noise, several robust algorithms have been developed.
Specifically, the lower-order statistics of the error signal were
utilized to combat impulsive interference, where the cost functions
included the mean absolute error \cite{Zheng2020Steady}, the mixed
error norm \cite{Zayyani2014Continuous}, etc. Some M-estimate
approaches were studied for the cost functions such as the Huber
functions to counteract the adverse effects of impulsive noise
\cite{Zhou2011New,Wang2020Constrained}. But such robust cost
functions cannot generally be smooth everywhere. Therefore, another
class of smoothly robust estimators was developed by using the
saturation property of error nonlinearities like the hyperbolic
tangent function \cite{Song2013Normalized}, and the maximum
correntropy criterion (MCC) \cite{Chen2014Steady}. The performance
of aforementioned EFLN-based algorithms would deteriorate under
impulsive interference. In \cite{Zhang2018Recursive}, a robust
recursive filtering algorithm based on EFLN and the Huber-norm was
proposed for nonlinear AEC against impulsive noise, but its
convergence analysis may be limited. Although the importance of the
robustness of the EFLN-based algorithm has been recognized, no
published works on robust EFLN-based filtering and its performance
analysis have been considered so far.

The above motivates this brief. In an attempt to further enhance the
performance of EFLN-based nonlinear filter in the presence of
impulsive interference, we introduce a new inverse square root (ISR)
cost function, which has smoothly saturating negative values and can
be more efficiently implemented \cite{Petersen2020Topological}.
Additionally, the theoretical analysis of the proposed EFLN-ISR
algorithm is strictly discussed in the mean-square sense. Finally,
the numerical and experimental studies are carried out to verify the
proposed EFLN-ISR algorithm.

\emph{Notations:} Notations in this brief are standard. At the time
index, the normal font letters with parentheses denote scalars, and
the lowercase boldface letters with parentheses denote column
vectors, respectively.

\section{System Model and EFLN-ISR Algorithm}

\subsection{System Model}

\begin{figure}[!ht]
\centering
\includegraphics[width=2.7in]{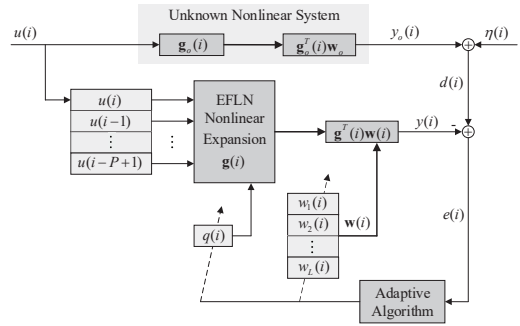}
\caption{EFLN-based nonlinear system identification model.}
\label{fig:1}
\end{figure}

Consider an adaptive filter based on EFLN to model a nonlinear
system, where the block diagram is depicted in Fig. \ref{fig:1},
whose structure consists of an EFLN-based nonlinear functional
expansion block followed by a linear filter. The input signal is
represented by $u(i)$ at a sample $i$.
$\mathbf{u}(i)=[u(i),u(i-1),\dots,u(i-P+1)]^T\in\mathbb{R}^{P\times1}$
arises as the input vector of a tapped delay line with the taps
number $P$. Using EFLN with the $N$-order functional expansion, the
$P$-dimension input vector is expanded to $L=P(2N+1)$ dimensions.
Referring to \cite{Patel2016Design,Patel2020Convergence}, the
expanded input vector is expressed as
\begin{align*}
\begin{split}
\mathbf{g}(i) = \big[\mathbf{g}_1^T(i),\mathbf{g}_2^T(i),\dots,\mathbf{g}_{2N+1}^T(i)\big]^T \in \mathbb{R}^{L \times 1}
\end{split}
\end{align*}
where its subvectors and the EFLN-based nonlinear expansion functions with the exponential factor $q(i)$ are listed in Table \ref{tab:1}. The filtered output signal is given by $y(i)=\mathbf{g}^T(i)\mathbf{w}(i)$, where $\mathbf{w}(i)=[w_1(i),w_2(i),\dots,w_L(i)]^T \in \mathbb{R}^{L \times 1}$ is the weights vector.

\begin{table}[!ht]\scriptsize
\centering
\renewcommand{\arraystretch}{1.2}
\setlength{\abovecaptionskip}{0pt}
\setlength{\belowcaptionskip}{0pt}
\caption{EFLN-based nonlinear expansion functions.}
\label{tab:1}
\begin{tabular}{ll}
\hline
$\mathbf{g}(i)$ &  Elements\\
\hline
  $\mathbf{g}_1^T(i)$  &   $u(i),u(i-1),\dots,u(i-P+1)$  \\
  $\mathbf{g}_2^T(i)$  &   $e^{-q(i)|u(i)|}\sin\big(\pi u(i)\big),\dots,e^{-q(i)|u(i-P+1)|}\sin\big(\pi u(i-P+1)\big)$  \\
  $\mathbf{g}_3^T(i)$  &   $e^{-q(i)|u(i)|}\cos\big(\pi u(i)\big),\dots,e^{-q(i)|u(i-P+1)|}\cos\big(\pi u(i-P+1)\big)$  \\
  \dots  &  \dots  \\
  $\mathbf{g}_{2N}^T(i)$  &   $e^{-q(i)|u(i)|}\sin\big(N\pi u(i)\big),\dots,e^{-q(i)|u(i-P+1)|}\sin\big(N\pi u(i-P+1)\big)$  \\
  $\mathbf{g}_{2N+1}^T(i)$  &   $e^{-q(i)|u(i)|}\cos\big(N\pi u(i)\big),\dots,e^{-q(i)|u(i-P+1)|}\cos\big(N\pi u(i-P+1)\big)$ \\
\hline
\end{tabular}
\end{table}

In the process of modeling, the noisy desired signal $d(i)$ is represented by
\begin{align*}
d(i)=y_o(i)+ \eta(i)= \mathbf{g}_o^T(i)\mathbf{w}_o + \eta(i)
\end{align*}
where $y_o(i) = \mathbf{g}_o^T(i)\mathbf{w}_o$ represents the system output signal, $\mathbf{g}_o(i)$ is the expanded input vector with the optimal exponential factor $q_o$, $\mathbf{w}_o \in \mathbb{R}^{L \times 1}$ is the optimal weights vector, and $\eta(i)$ is a zero-mean additive noise. Additionally, the \emph{a priori error} $\xi(i)$ and the error signal $e(i)$ are respectively defined as
\begin{align*}
\begin{split}
\xi(i) &= y_o(i)-y(i) = \mathbf{g}_o^T(i)\mathbf{w}_o - \mathbf{g}^T(i)\mathbf{w}(i) \\
e(i) &= d(i) - y(i) = \xi(i) + \eta(i)  .
\end{split}
\end{align*}

\subsection{Proposed EFLN-ISR Algorithm}

The existing EFLN-based filters always minimize $e^2(i)$ to get the
corresponding EFLN-LMS algorithms, but they may perform poorly under
impulsive interference. To make the EFLN-based algorithm with
preferable robustness, we thus define a new ISR cost function as
follows
\begin{align}\label{eq:costf}
Q\big(e(i)\big)={ \textstyle \frac{1}{2} }e^2(i)[1+\lambda e^4(i)]^{-\frac{1}{2}}
\end{align}
where $\lambda >0$ is a scalar parameter. Taking the derivative of \eqref{eq:costf} results in
\begin{align*}
r\big(e(i)\big) = {\textstyle \frac{\partial Q\left (e(i)\right )}{\partial e(i)} }= e(i)[1+\lambda e^4(i)]^{-\frac{3}{2}}.
\end{align*}

\begin{figure}[!ht]
\centering
\includegraphics[width=3in]{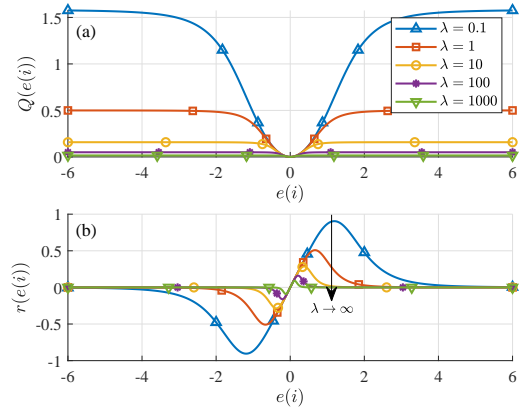}
\caption{(a) Curves of the ISR cost function with different $\lambda$. (b) Slope curves of the ISR cost function with different $\lambda = \{0.1, 1, 10, 100, 1000\}$.}
\label{fig:2}
\end{figure}

The curves of $Q\big(e(i)\big)$ and its derivative $r\big(e(i)\big)$ with different $\lambda$ are described in Fig. \ref{fig:2}, which show that the evolutions of $Q\big(e(i)\big)$ are concave around zero and flat away from it. Clearly, the slope curves approach zero with increasing errors, which will reduce the sensitivity to large outliers. When $\lambda$ is larger, the evaluations of slope curves away from zero will be smaller, which demonstrates strong robustness against impulsive interference. So the appropriate $\lambda$ plays a significant role for the robustness and performance.

Taking advantage of the gradient descent criterion to derive the proposed EFLN-ISR algorithm, the learning rules of the weights vector and the exponential factor are updated as
\begin{align*}
\mathbf{w}(i+1) &= \mathbf{w}(i) - \mu_w {\textstyle \frac{\partial Q\left (e(i)\right )}{\partial \mathbf{w}(i) } }\\
q(i+1) &= q(i) - \mu_q {\textstyle \frac{\partial Q\left (e(i)\right )}{\partial q(i) }  }
\end{align*}
where $\mu_w$ and $\mu_q$ denote the step sizes. The gradients with respect to $\mathbf{w}(i)$ and $q(i)$ can be calculated by the derivative chain rule, and thus the iterative learning rules of the EFLN-ISR algorithm can be derived as
\begin{align}
\mathbf{w}(i+1) &= \mathbf{w}(i) + \mu_w e(i)\big[1+\lambda e^4(i)\big]^{-\frac{3}{2}} \mathbf{g}(i)\label{eq:w}\\
q(i+1) &= q(i) + \mu_q e(i)\big[1+\lambda e^4(i)\big]^{-\frac{3}{2}} \mathbf{h}^T(i)\mathbf{w}(i) \label{eq:q}
\end{align}
where $\mathbf{h}(i)= {\textstyle \frac{\partial \mathbf{g}(i)}{\partial q(i)}=\Big[\frac{\partial \mathbf{g}_1^T(i)}{\partial q(i)},\frac{\partial \mathbf{g}_2^T(i)}{\partial q(i)},\dots,\frac{\partial \mathbf{g}_{2N+1}^T(i)}{\partial q(i)} \Big]^T\in \mathbb{R}^{L \times 1} }$ with its elements being listed in Table \ref{tab:2}.
\begin{table}[!ht]\normalsize
\centering
\renewcommand{\arraystretch}{1.2}
\setlength{\abovecaptionskip}{0pt}
\setlength{\belowcaptionskip}{0pt}
\caption{Elements of $\mathbf{h}(i)$.}
\label{tab:2}
\resizebox{\linewidth}{!}{
\begin{tabular}{ll}
\hline
$\mathbf{h}(i)$ &  Elements\\
\hline
  $\partial \mathbf{g}_{1}^T(i)/\partial q(i)$  &   $0,0,\dots,0$  \\
  $\partial \mathbf{g}_{2}^T(i)/\partial q(i)$  &   $-|u(i)|e^{-q(i)|u(i)|}\sin\big(\pi u(i)\big),\dots,-|u(i-P+1)|e^{-q(i)|u(i-P+1)|}\sin\big(\pi u(i-P+1)\big)$  \\
  $\partial \mathbf{g}_{3}^T(i)/\partial q(i)$  &   $-|u(i)|e^{-q(i)|u(i)|}\cos\big(\pi u(i)\big),\dots,-|u(i-P+1)|e^{-q(i)|u(i-P+1)|}\cos\big(\pi u(i-P+1)\big)$  \\
  \dots  &  \dots  \\
  $\partial \mathbf{g}_{2N}^T(i)/\partial q(i)$  &   $-|u(i)|e^{-q(i)|u(i)|}\sin\big(N\pi u(i)\big),\dots,-|u(i-P+1)|e^{-q(i)|u(i-P+1)|}\sin\big(N\pi u(i-P+1)\big)$  \\
  $\partial \mathbf{g}_{2N+1}^T(i)/\partial q(i)$  &   $-|u(i)|e^{-q(i)|u(i)|}\cos\big(N\pi u(i)\big),\dots,-|u(i-P+1)|e^{-q(i)|u(i-P+1)|}\cos\big(N\pi u(i-P+1)\big)$ \\
\hline
\end{tabular} }
\end{table}

\section{Performance Analysis}

We present the theoretical analysis of the EFLN-ISR algorithm in this section. The following assumptions are made throughout this brief.
\begin{assumption}\label{as:iid}
The zero-mean additive noise $\eta(i)$ is independent of $\mathbf{u}(i),\mathbf{w}(i)$ and $q(i)$.
\end{assumption}
\begin{assumption}\label{as:each}
$\mathbf{u}(i), \mathbf{w}(i)$ and $q(i)$ are mutually statistically independent.
\end{assumption}
\begin{assumption}\label{as:priori}
$\xi(i)$ is asymptotically uncorrelated with $||\mathbf{g}(i)||^2$ and $|\mathbf{h}^T(i)\mathbf{w}(i)|^2$ at steady-state.
\end{assumption}

According to \eqref{eq:w}, \eqref{eq:q} and some mathematical operations, as well as Assumptions \ref{as:iid} and \ref{as:each}, we have
\begin{align*}
\lim_{i\rightarrow\infty}\mathbb{E}\{\mathbf{w}(i)\} = \mathbf{w}_o~\mathrm{and}~\lim_{i\rightarrow\infty}\mathbb{E}\{q(i)\} = q_o
\end{align*}
at steady-state. The detailed derivations for this mean analysis are provided in supplementary material.

As a performance metric in the mean-square sense, the steady-state excess mean-square error (EMSE) is defined as
\begin{align*}
\begin{split}
\mathrm{EMSE}(\infty)=\lim_{i\rightarrow\infty}\mathbb{E}\{\xi^2(i)\}=\lim_{i\rightarrow\infty}\mathbb{E}\Big\{\big[\mathbf{g}_o^T(i)\tilde {\mathbf{w}}(i)+ \tilde q(i) \mathbf{h}^T(i)\mathbf{w}(i)\big]^2\Big\}
\end{split}
\end{align*}
where $\tilde {\mathbf{w}}(i) = \mathbf{w}_o-\mathbf{w}(i)$ denotes the weights vector error, and $\tilde q(i) = q_o-q(i)$ denotes the exponential factor error, respectively. It is significant that we focus on the steady-state EMSE, and let $\xi_w(i)=\mathbf{g}_o^T(i)\tilde {\mathbf{w}}(i)$ and $\xi_q(i)=\tilde q(i) \mathbf{h}^T(i)\mathbf{w}(i)$. In the following, the theoretical evaluation of $\mathrm{EMSE}(\infty)$, i.e., $\lim_{i\rightarrow\infty}\mathbb{E}\{\xi^2(i)\}$ will be derived.

\subsection{Mean-Square Performance of $\mathbf{w}(i)$}

Inserting \eqref{eq:w}, the weights vector error is evaluated as
\begin{align}\label{eq:we}
\begin{split}
\tilde {\mathbf{w}}(i+1) =\tilde {\mathbf{w}}(i)-\mu_w r(e(i))\mathbf{g}(i)  .
\end{split}
\end{align}
Calculating the energy of two sides of \eqref{eq:we} and making the expectation operation for the energy relation, one has
\begin{align*}
\begin{split}
\mathbb{E}\big\{||\tilde {\mathbf{w}}(i+1)||^2\big\} =&~\mathbb{E}\big\{||\tilde {\mathbf{w}}(i)||^2\big\}-2\mu_w\mathbb{E}\big\{r(e(i))\mathbf{g}^T(i)\tilde {\mathbf{w}}(i)\big\}\\
&~+\mu_w^2\mathbb{E} \big\{r^2(e(i))||\mathbf{g}(i)||^2\big\}   .
\end{split}
\end{align*}
It is noting that $\lim_{i\rightarrow\infty}\mathbb{E}\big\{||\tilde {\mathbf{w}}(i+1)||^2\big\} = \lim_{i\rightarrow\infty}\mathbb{E}\big\{||\tilde {\mathbf{w}}(i)||^2\big\}$ holds, thus resulting in
\begin{align}\label{eq:wss}
\begin{split}
2\mathbb{E}\big\{r(e(i))\mathbf{g}^T(i)\tilde {\mathbf{w}}(i)\big\}=\mu_w\mathbb{E} \big\{r^2(e(i))||\mathbf{g}(i)||^2\big\}   .
\end{split}
\end{align}
At steady-state for $i\rightarrow\infty$, assuming that $q(i)\rightarrow q_o$, then $\mathbf{g}(i) \rightarrow \mathbf{g}_o(i)$, we have $e(i)\approx\xi_w(i) + \eta(i)$. The following Taylor expansion of $r(e(i))$ has been taken as
\begin{align*}
\begin{split}
r(e(i)) = r(\eta(i))+r'(\eta(i))\xi_w(i)+{ \textstyle \frac{1}{2}  } r''(\eta(i))\xi_w^2(i)+ h.o.t  .
\end{split}
\end{align*}
Using Assumption \ref{as:iid} and considering the above Taylor expansion, the left side of \eqref{eq:wss} results in
\begin{align}\label{eq:wls}
\begin{split}
2\mathbb{E}\big\{r(e(i))\mathbf{g}^T(i)\tilde {\mathbf{w}}(i)\big\}= 2\mathbb{E}\{r'(\eta(i))\}\mathbb{E}\{\xi_w^2(i)\}  .
\end{split}
\end{align}
By Assumption \ref{as:priori}, the right side of \eqref{eq:wss} can be calculated as
\begin{align}\label{eq:wrs}
\begin{split}
&~\mu_w\mathbb{E} \big\{r^2(e(i))||\mathbf{g}(i)||^2\big\} \\
=&~\mu_w \mathbb{E} \big\{ r^2(\eta(i))\big\}\mathbb{E} \big\{||\mathbf{g}(i)||^2\big\}+\mu_w \big [ \mathbb{E} \big\{ r'^2(\eta(i))\big\}\\
&~+\mathbb{E} \big\{ r(\eta(i))r''(\eta(i))\big\}\big]\mathbb{E} \big\{||\mathbf{g}(i)||^2\big\}\mathbb{E} \big\{\xi_w^2(i)\big\}   .
\end{split}
\end{align}
Inserting \eqref{eq:wls} and \eqref{eq:wrs} and concerning the steady-state for $i\rightarrow\infty$, we can obtain
\begin{align}\label{eq:wssv}
\begin{split}
\lim_{i\rightarrow\infty}\mathbb{E} \big\{\xi_w^2(i)\big\} = \frac{ \mu_w  E_1 \mathbb{E} \{||\mathbf{g}(i)||^2\} }{ 2 E_2 -\mu_w E_3 \mathbb{E} \{||\mathbf{g}(i)||^2\} }
\end{split}
\end{align}
where we denote as $E_1 \triangleq \mathbb{E} \{ r^2(\eta(i)) \} = \mathbb{E}\{\eta^2(i)[1+\lambda \eta^4(i)]^{-3}\}$, $E_2 \triangleq \mathbb{E} \{ r' (\eta(i)) \}=\mathbb{E}\{ [1-5\lambda \eta^4(i)][1+\lambda \eta^4(i)]^{-\frac{5}{2}} \}$, and $E_3 \triangleq  \mathbb{E}  \{ r'^2(\eta(i)) \} +\mathbb{E} \{ r(\eta(i))r''(\eta(i)) \}=\mathbb{E} \{ [1-5\lambda \eta^4(i)]^2[1+\lambda \eta^4(i)]^{-5}   \} +\mathbb{E} \{ -30\lambda \eta^4(i)[1-\lambda \eta^4(i)]   [1+\lambda \eta^4(i)]^{-5}\}$.

\subsection{Mean-Square Performance of $q(n)$}

Similarly, subtracting both sides of \eqref{eq:q} from $q_o$ yields
\begin{align}\label{eq:qe}
\begin{split}
\tilde q (i+1) =\tilde q (i)-\mu_q r(e(i))\mathbf{h}^T(i)\mathbf{w}(i)  .
\end{split}
\end{align}
Evaluating the energy of two sides of \eqref{eq:qe} and taking expectation, and noting that $\mathbb{E}\big\{\tilde q ^2(i+1)\big\} = \mathbb{E}\big\{\tilde q^2(i)\big\}$ at steady-state, it yields
\begin{align}\label{eq:qss}
\begin{split}
2\mathbb{E}\big\{r(e(i))\xi_q (i) \big\}=\mu_q \mathbb{E} \big\{r^2(e(i))|\mathbf{h}^T(i)\mathbf{w}(i)|^2 \big\} .
\end{split}
\end{align}
Assuming in this phase that $\mathbf{w}(i)\rightarrow \mathbf{w}_o$ as $i\rightarrow\infty$, and thus we have $e(i)\approx\xi_q(i) + \eta(i)$. Then, according to the above discussions, the both sides of \eqref{eq:qss} are calculated as
\begin{align}\label{eq:qls}
\begin{split}
2\mathbb{E}\big\{r(e(i))\xi_q(i)\big\}\approx 2\mathbb{E}\{r'(\eta(i))\}\mathbb{E}\{\xi_q^2(i)\}
\end{split}
\end{align}
and
\begin{align}\label{eq:qrs}
\begin{split}
&~\mu_q\mathbb{E} \big\{r^2(e(i))|\mathbf{h}^T(i)\mathbf{w}(i)|^2\big\} \\
=&~\mu_q \mathbb{E} \big\{ r^2(\eta(i))\big\}\mathbb{E} \big\{|\mathbf{h}^T(i)\mathbf{w}(i)|^2\big\}+\mu_q \Big [ \mathbb{E} \big\{ r'^2(\eta(i))\big\}\\
&~+\mathbb{E} \big\{ r(\eta(i))r''(\eta(i))\big\}\Big]\mathbb{E} \big\{|\mathbf{h}^T(i)\mathbf{w}(i)|^2\big\}\mathbb{E} \big\{\xi_q^2(i)\big\}  .
\end{split}
\end{align}

Substituting \eqref{eq:qls} and \eqref{eq:qrs} into \eqref{eq:qss} and evaluating them at steady-state for $i\rightarrow\infty$, we obtain
\begin{align}\label{eq:qssv}
\begin{split}
\lim_{i\rightarrow\infty}\mathbb{E} \big\{\xi_q^2(i)\big\} = \frac{ \mu_w  E_1 \mathbb{E} \big\{|\mathbf{h}^T(i)\mathbf{w}(i)|^2\big\}  }{ 2 E_2 -\mu_w E_3 \mathbb{E} \big\{|\mathbf{h}^T(i)\mathbf{w}(i)|^2\big\}  }  .
\end{split}
\end{align}

\subsection{Steady-State EMSE}
Taking into account the definition of steady-state EMSE, we can derive as
\begin{align*}
\begin{split}
\mathrm{EMSE}(\infty)&=\lim_{i\rightarrow\infty}\mathbb{E}\big\{\big[\xi_w(i)+\xi_q(i)\big]^2\big\} \\
&=\lim_{i\rightarrow\infty}\mathbb{E} \big\{\xi_w^2(i)\big\} +2\lim_{i\rightarrow\infty}\mathbb{E} \big\{\xi_w(i)\xi_q(i)\big\} +\lim_{i\rightarrow\infty}\mathbb{E} \big\{\xi_q^2(i)\big\}  .
\end{split}
\end{align*}
Since $\mathbf{w}(i) \rightarrow \mathbf{w}_o$ and $q(i) \rightarrow q_o$ at steady-state, it implies that
\begin{align*}
\begin{split}
\lim_{i\rightarrow\infty}\mathbb{E} \big\{\xi_w(i)\xi_q(i)\big\}=\lim_{i\rightarrow\infty}\mathrm{Tr}\big[\mathbb{E}\{\tilde {\mathbf{w}}(i)\mathbf{w}^T(i)\} \mathbb{E}\{\tilde q(i)\mathbf{h}(i)\mathbf{g}_o^T(i)\} \big]=0   .
\end{split}
\end{align*}
Thus, the steady-state EMSE is
\begin{align*}
\begin{split}
\mathrm{EMSE}(\infty) =\lim_{i\rightarrow\infty}\mathbb{E} \big\{\xi_w^2(i)\big\} +\lim_{i\rightarrow\infty}\mathbb{E} \big\{\xi_q^2(i)\big\}
\end{split}
\end{align*}
where this theoretical value can be provided by \eqref{eq:wssv} and \eqref{eq:qssv}.

\section{Simulation and Experimental Studies}

\subsection{Case 1: Verification of Analysis}

Consider an unknown nonlinear system based on EFLN expansion, which is given by $\mathbf{g}_o(i)=\big[u(i),e^{-q_o|u(i)|}\sin\big(\pi u(i)\big),$ $e^{-q_o|u(i)|}\cos\big(\pi u(i)\big),~e^{-q_o|u(i)|}\sin\big(2\pi u(i)\big),~e^{-q_o|u(i)|}\cos\big(2\pi u(i)\big),$ $u(i-1),~e^{-q_o|u(i-1)|}\sin\big(\pi u(i-1)\big),~e^{-q_o|u(i-1)|}\cos\big(\pi u(i-1)\big),$ $e^{-q_o|u(i-1)|}\sin\big(2\pi u(i-1)\big),e^{-q_o|u(i-1)|}\cos\big(2\pi u(i-1)\big) \big]^T$, and the optimal weight vector and the exponential factor are given by $\mathbf{w}_o=[0.3,0.6,-0.2,0.05,-0.27,-0.3,0.4,-0.1,0.01,0.25]^T$ and $q_o=-0.2$, respectively.

\begin{figure}[!ht]
\centering
\includegraphics[width=3.4in]{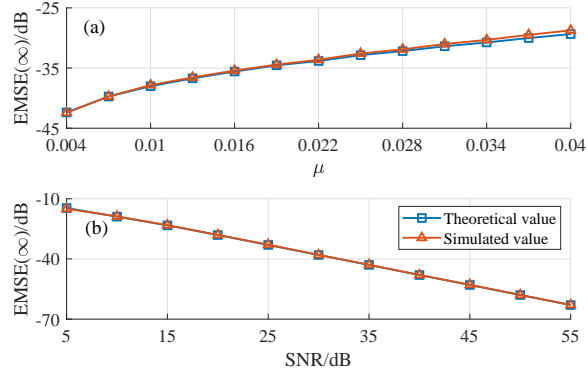}
\caption{Theoretical and simulated $\mathrm{EMSE}(\infty)$ under Gaussian interference. (a) $\mathrm{EMSE}(\infty)$ versus step sizes $\mu_w=\mu_q=\{0.004,0.007,\dots,0.04\}$, with SNR = 30dB. (b) $\mathrm{EMSE}(\infty)$ versus SNRs $=\{5,10,\dots,55\}$dB, with $\mu_w=\mu_q=0.01$.}
\label{fig:3}
\end{figure}

In this study, $u(i)$ is extracted from the Gaussian signal with the mean 0 and the variance 1. Fig. \ref{fig:3} checks the analysis results of EFLN-ISR under Gaussian interference. The expectations involved in the theoretical steady-state value, given by \eqref{eq:wssv} and \eqref{eq:qssv}, are calculated as the average of the final 1,000 samples over 50 independent trials. The simulated value is also obtained by averaging over 50 independent runs. As can be seen, a good agreement compared with the theoretical and simulated findings is quite apparent.

\subsection{Case 2: Under Impulsive Interference}

In this case, we consider a nonlinear system describing the asymmetric loudspeaker distortion \cite{Patel2016Design,Patel2020Convergence}, whose input-output relation is expressed as $y_o(i) = \beta \big [1/\big( 1+e^{-\rho \kappa(i)} \big)-0.5  \big ]$, where $\beta =2$ is the system gain, and $\rho$ represents the slope parameter given by $\rho=4$ if $\kappa(i)>0$ and $\rho=0.5$ if $\kappa(i)\le 0$ with $\kappa(i)=1.5u(i)-0.3u^2(i)$. The input $u(i)$ is employed by a uniformly distributed signal over the interval $[-0.5,0.5]$. We consider that the $\alpha$-stable distribution is used to characterize the additive noise with its impulsive nature, whose characteristic function is given by $\psi(t) = e^{-\gamma^\alpha|t|^\alpha}$, where the characteristic parameter is $0< \alpha \le 2$ with the minor $\alpha$ resulting in more outliers, and the dispersion parameter is $\gamma>0$.

\begin{figure}[!ht]
\centering
\includegraphics[width=3in]{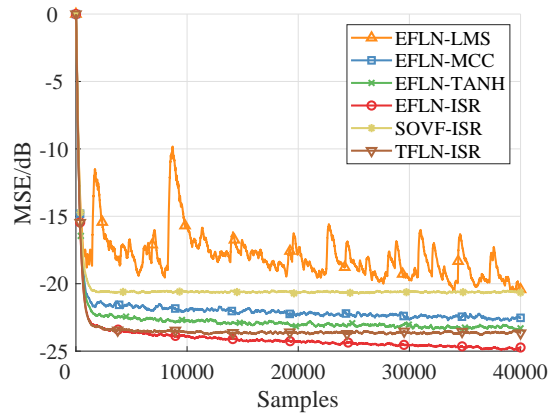}
\caption{Comparison of different nonlinear filtering algorithms under impulsive interference, with $\alpha=1.6,\gamma=0.05$ and SNR = 30dB.}
\label{fig:4}
\end{figure}
\begin{figure}[!ht]
\centering
\includegraphics[width=3in]{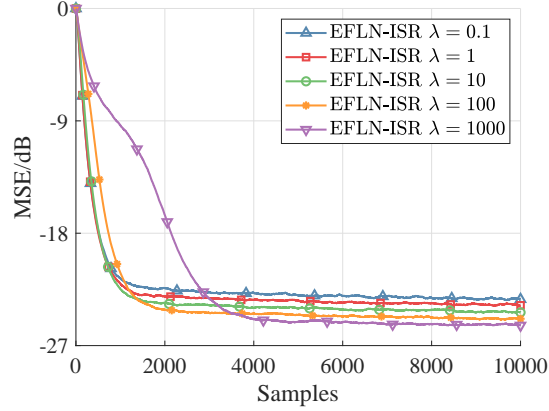}
\caption{Comparison of EFLN-ISR with different $\lambda = \{0.1, 1, 10, 100, 1000\}$ under impulsive interference, with $\alpha=1.6,\gamma=0.05$ and SNR = 30dB.}
\label{fig:5}
\end{figure}
In the nonlinear system identification task, the performance of EFLN-ISR is compared with that obtained using the EFLN-LMS, EFLN based on maximum correntropy criterion (EFLN-MCC), EFLN based on hyperbolic tangent function (EFLN-TANH), SOVF based on ISR (SOVF-ISR) and TFLN based on ISR (TFLN-ISR) algorithms. For a fair comparison, the simulation parameters of all algorithms are taken to hold the same initial convergence. Fig. \ref{fig:4} shows the results of nonlinear filtering algorithms against impulsive interference. It is clear that the proposed EFLN-ISR algorithm has much smaller steady-state error than others under the same initial convergence. In addition, the performance of EFLN-ISR with different $\lambda$ is investigated in Fig. \ref{fig:5}. We can also find that if the parameter $\lambda$ is smaller, the convergence will be faster but it increases the steady-state error. Nevertheless, when $\lambda$ is too large, it results in the slower convergence.

\subsection{Case 3: Hysteresis System Identification}

\begin{figure}[!ht]
\centering
\includegraphics[width=3.2in]{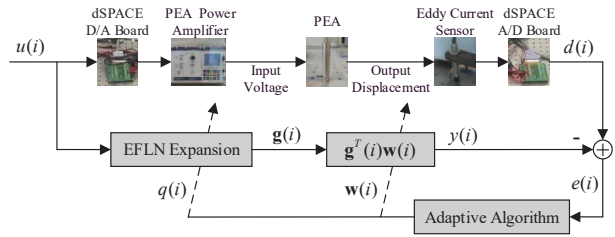}
\caption{Block diagram of the nonlinear model identification of PEA.}
\label{fig:6}
\end{figure}
To certify the validity of the proposed EFLN-ISR algorithm, EFLN-ISR has been applied to the practical hysteresis system identification. As shown in Fig. \ref{fig:6}, the use of EFLN-ISR to model the piezoelectric actuator (PEA), whose input-output relation has hysteretic effect, is implemented in the experimental platform controlled by the dSPACE system. In this experiment, a type of PSt150/7/60VS12 PEA is utilized for the identified plant, which can move the 58.83\textmu m displacements. A dSPACE DS1006 processor board is adopted to transfer the input voltage and output displacement between the PEA and the computer. The input voltages are processed by a dSPACE DS2103 board provided with 32$\times$14-bits D/A converter channels, and further driven by a PEA servo power amplifier. The output displacements are measured by an eddy current sensor with 8mV/\textmu m resolution ratio, and recorded by a dSPACE DS2002 board provided with 32$\times$16-bits A/D converter channels. The sampling frequency of the whole experimental process is taken as 10kHz, and the dSPACE ControlDesk environment visually presents the experimental results in real time.

\begin{figure}[!ht]
\centering
\includegraphics[width=3.3in]{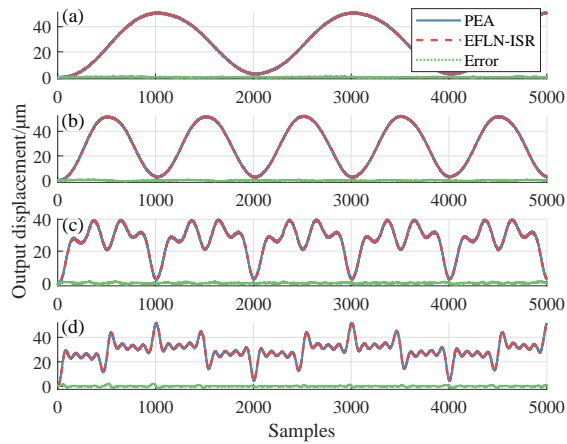}
\caption{Comparison of the output displacement measured by experiment data and that estimated by the EFLN-ISR algorithm. (a) With the single frequency of 5Hz. (b) With the single frequency of 10Hz. (c) With the complex frequency of 10/20/40Hz. (d) With the complex frequency of 5/25/45/65/85Hz.}
\label{fig:7}
\end{figure}

We set a sinusoidal signal $u(i)$ as the input voltage of PEA. The hysteresis model of PEA is identified by the EFLN-ISR and SOVF-ISR algorithms, whose parameters are chosen as $P=7,N=2,\mu_w=\mu_q=0.02$ and $\lambda=1$. To further evaluate the identification precision of adaptive algorithms, the sinusoidal signals with the single and complex frequency have been taken as the input voltages of PEA. Moreover, the root mean-square error $\mathrm{RMSE} = \scriptstyle \sqrt{\sum_{i=1}^M e^2(i)/M}$, the relative error $\mathrm{RE} = {\scriptstyle \sqrt{\sum_{i=1}^M e^2(i)/\sum_{i=1}^M d^2(i)}  }\times 100\%$ and the maximum absolute error $\mathrm{MAE}=\max |e(i)|$ with $M$ sample data are used to conduct the quantitative evaluation. Fig. \ref{fig:7} shows the comparison of the experimental data and filtered outputs with the input signal of the single and complex frequency. Results of identification precision for the EFLN-ISR and SOVF-ISR algorithms are listed in Table \ref{tab:3}. From experimental results through the proposed EFLN-ISR algorithm, we can see that the measured displacement and the filtered output are almost the same, and the RMSE, RE and MAE indicators are reduced as compared with the SOVF-ISR algorithm.

\begin{table}[!ht]\scriptsize
\renewcommand{\arraystretch}{1.2}
\caption{Identification Precision of the EFLN-ISR and SOVF-ISR Algorithms}
\label{tab:3}
\centering
\begin{tabular}{|c|c|c|c|c|c|c|}
\hline
\multicolumn{1}{|c|}{\multirow{2}{*}{ \makecell{Input \\Frequency/Hz} } } & \multicolumn{2}{c|}{RMSE/\textmu m}    & \multicolumn{2}{c|}{RE}    & \multicolumn{2}{c|}{MAE/\textmu m}  \\
\cline{2-7}
\multicolumn{1}{|c|}{}       & \multicolumn{1}{c|}{EFLN} & \multicolumn{1}{c|}{SOVF} & \multicolumn{1}{c|}{EFLN}& \multicolumn{1}{c|}{SOVF} & \multicolumn{1}{c|}{EFLN}& \multicolumn{1}{c|}{SOVF}     \\
\hline
5               & 0.4721           & 0.4836          & 1.47\%          & 1.51\%         & 1.4450          & 1.5951      \\
\hline
10              & 0.5267           & 0.5680          & 1.59\%          & 1.71\%         & 1.4625          & 1.9246      \\
\hline
10/20/40        & 0.5580           & 0.6742          & 1.97\%          & 2.38\%         & 1.8423          & 3.0757      \\
\hline
5/25/45/65/85   & 0.6754           & 0.7406          & 2.23\%          & 2.45\%         & 2.9915          & 5.5706      \\
\hline
\end{tabular}
\end{table}

\section{Conclusion}

This brief proposed an EFLN-based nonlinear filtering algorithm derived by minimizing a new ISR cost function, which possesses preferable robustness under impulsive interference. The steady-state performance analysis was rigorously given. Simulation studies validate the performance findings. Besides, the validity of the proposed EFLN-ISR algorithm is confirmed by the practical results of hysteresis system identification.

\bibliographystyle{IEEEtran}
\bibliography{IEEEref}

\begin{thebibliography}{10}
\providecommand{\url}[1]{#1}
\csname url@samestyle\endcsname
\providecommand{\newblock}{\relax}
\providecommand{\bibinfo}[2]{#2}
\providecommand{\BIBentrySTDinterwordspacing}{\spaceskip=0pt\relax}
\providecommand{\BIBentryALTinterwordstretchfactor}{4}
\providecommand{\BIBentryALTinterwordspacing}{\spaceskip=\fontdimen2\font plus
\BIBentryALTinterwordstretchfactor\fontdimen3\font minus
  \fontdimen4\font\relax}
\providecommand{\BIBforeignlanguage}[2]{{%
\expandafter\ifx\csname l@#1\endcsname\relax
\typeout{** WARNING: IEEEtran.bst: No hyphenation pattern has been}%
\typeout{** loaded for the language `#1'. Using the pattern for}%
\typeout{** the default language instead.}%
\else
\language=\csname l@#1\endcsname
\fi
#2}}
\providecommand{\BIBdecl}{\relax}
\BIBdecl

\bibitem{Comminiello2018Adaptive}
D.~Comminiello and J.~C. Pr\'{\i}ncipe, \emph{Adaptive Learning Methods for
  Nonlinear System Modeling}.\hskip 1em plus 0.5em minus 0.4em\relax
  Kidlington, Oxford, UK: Elsevier, 2018.

\bibitem{jidf}
R.~C. {de Lamare} and R.~{Sampaio-Neto}, ``Adaptive reduced-rank processing
  based on joint and iterative interpolation, decimation, and filtering,''
  \emph{IEEE Transactions on Signal Processing}, vol.~57, no.~7, pp.
  2503--2514, 2009.

\bibitem{jidf_echo}
M.~{Yukawa}, R.~C. {de Lamare}, and R.~{Sampaio-Neto}, ``Efficient acoustic
  echo cancellation with reduced-rank adaptive filtering based on selective
  decimation and adaptive interpolation,'' \emph{IEEE Transactions on Audio,
  Speech, and Language Processing}, vol.~16, no.~4, pp. 696--710, 2008.

\bibitem{ccg}
\BIBentryALTinterwordspacing
L.~Wang, ``\BIBforeignlanguage{English}{Constrained adaptive filtering
  algorithms based on conjugate gradient techniques for beamforming},''
  \emph{\BIBforeignlanguage{English}{IET Signal Processing}}, vol.~4, pp.
  686--697(11), December 2010. [Online]. Available:
  \url{https://digital-library.theiet.org/content/journals/10.1049/iet-spr.2009.0243}
\BIBentrySTDinterwordspacing

\bibitem{jio}
R.~C. {de Lamare} and R.~{Sampaio-Neto}, ``Reduced-rank adaptive filtering
  based on joint iterative optimization of adaptive filters,'' \emph{IEEE
  Signal Processing Letters}, vol.~14, no.~12, pp. 980--983, 2007.

\bibitem{intadap}
------, ``Adaptive reduced-rank mmse filtering with interpolated fir filters
  and adaptive interpolators,'' \emph{IEEE Signal Processing Letters}, vol.~12,
  no.~3, pp. 177--180, 2005.

\bibitem{saalt}
------, ``Sparsity-aware adaptive algorithms based on alternating optimization
  and shrinkage,'' \emph{IEEE Signal Processing Letters}, vol.~21, no.~2, pp.
  225--229, 2014.

\bibitem{jiols}
------, ``Reduced--rank space--time adaptive interference suppression with
  joint iterative least squares algorithms for spread--spectrum systems,''
  \emph{IEEE Transactions on Vehicular Technology}, vol.~59, no.~3, pp.
  1217--1228, 2010.

\bibitem{ccmjio}
L.~{Wang}, R.~C. {de Lamare}, and M.~{Yukawa}, ``Adaptive reduced-rank
  constrained constant modulus algorithms based on joint iterative optimization
  of filters for beamforming,'' \emph{IEEE Transactions on Signal Processing},
  vol.~58, no.~6, pp. 2983--2997, 2010.

\bibitem{wlmwf}
N.~{Song}, R.~C. {de Lamare}, M.~{Haardt}, and M.~{Wolf}, ``Adaptive widely
  linear reduced-rank interference suppression based on the multistage wiener
  filter,'' \emph{IEEE Transactions on Signal Processing}, vol.~60, no.~8, pp.
  4003--4016, 2012.

\bibitem{wljio}
N.~{Song}, W.~U. {Alokozai}, R.~C. {de Lamare}, and M.~{Haardt}, ``Adaptive
  widely linear reduced-rank beamforming based on joint iterative
  optimization,'' \emph{IEEE Signal Processing Letters}, vol.~21, no.~3, pp.
  265--269, 2014.

\bibitem{jiomimo}
R.~C. {de Lamare} and R.~{Sampaio-Neto}, ``Adaptive reduced-rank equalization
  algorithms based on alternating optimization design techniques for mimo
  systems,'' \emph{IEEE Transactions on Vehicular Technology}, vol.~60, no.~6,
  pp. 2482--2494, 2011.

\bibitem{sjidf}
R.~{Fa}, R.~C. {de Lamare}, and L.~{Wang}, ``Reduced-rank stap schemes for
  airborne radar based on switched joint interpolation, decimation and
  filtering algorithm,'' \emph{IEEE Transactions on Signal Processing},
  vol.~58, no.~8, pp. 4182--4194, 2010.

\bibitem{saabf}
S.~{Li}, R.~C. {de Lamare}, and R.~{Fa}, ``Reduced-rank linear interference
  suppression for ds-uwb systems based on switched approximations of adaptive
  basis functions,'' \emph{IEEE Transactions on Vehicular Technology}, vol.~60,
  no.~2, pp. 485--497, 2011.

\bibitem{rrser}
Y.~{Cai}, R.~C. {de Lamare}, B.~{Champagne}, B.~{Qin}, and M.~{Zhao},
  ``Adaptive reduced-rank receive processing based on minimum symbol-error-rate
  criterion for large-scale multiple-antenna systems,'' \emph{IEEE Transactions
  on Communications}, vol.~63, no.~11, pp. 4185--4201, 2015.

\bibitem{jiostap}
R.~{Fa} and R.~C. {De Lamare}, ``Reduced-rank stap algorithms using joint
  iterative optimization of filters,'' \emph{IEEE Transactions on Aerospace and
  Electronic Systems}, vol.~47, no.~3, pp. 1668--1684, 2011.

\bibitem{locsme}
H.~{Ruan} and R.~C. {de Lamare}, ``Robust adaptive beamforming using a
  low-complexity shrinkage-based mismatch estimation algorithm,'' \emph{IEEE
  Signal Processing Letters}, vol.~21, no.~1, pp. 60--64, 2014.

\bibitem{oskpme}
------, ``Robust adaptive beamforming based on low-rank and cross-correlation
  techniques,'' \emph{IEEE Transactions on Signal Processing}, vol.~64, no.~15,
  pp. 3919--3932, 2016.

\bibitem{dce}
S.~{Xu}, R.~C. {de Lamare}, and H.~V. {Poor}, ``Distributed compressed
  estimation based on compressive sensing,'' \emph{IEEE Signal Processing
  Letters}, vol.~22, no.~9, pp. 1311--1315, 2015.

\bibitem{mskaesprit}
S.~F.~B. {Pinto} and R.~C. {de Lamare}, ``Multistep knowledge-aided iterative
  esprit: Design and analysis,'' \emph{IEEE Transactions on Aerospace and
  Electronic Systems}, vol.~54, no.~5, pp. 2189--2201, 2018.

\bibitem{damdc}
T.~G. {Miller}, S.~{Xu}, R.~C. {de Lamare}, and H.~V. {Poor}, ``Distributed
  spectrum estimation based on alternating mixed discrete-continuous
  adaptation,'' \emph{IEEE Signal Processing Letters}, vol.~23, no.~4, pp.
  551--555, 2016.

\bibitem{rdrls}
Y.~{Yu}, H.~{Zhao}, R.~C. {de Lamare}, Y.~{Zakharov}, and L.~{Lu}, ``Robust
  distributed diffusion recursive least squares algorithms with side
  information for adaptive networks,'' \emph{IEEE Transactions on Signal
  Processing}, vol.~67, no.~6, pp. 1566--1581, 2019.

\bibitem{dlmm}
Y.~{Yu}, H.~{He}, T.~{Yang}, X.~{Wang}, and R.~C. {de Lamare}, ``Diffusion
  normalized least mean m-estimate algorithms: Design and performance
  analysis,'' \emph{IEEE Transactions on Signal Processing}, vol.~68, pp.
  2199--2214, 2020.

\bibitem{Sicuranza2012BIBO}
G.~L. Sicuranza and A.~Carini, ``On the {BIBO} stability condition of adaptive
  recursive {FLANN} filters with application to nonlinear active noise
  control,'' \emph{IEEE Trans. Audio, Speech, Lang. Process.}, vol.~20, no.~1,
  pp. 234--245, Jan. 2012.

\bibitem{Tan2001Adaptive}
L.~Tan and J.~Jiang, ``Adaptive {Volterra} filters for active control of
  nonlinear noise processes,'' \emph{IEEE Trans. Signal Process.}, vol.~49,
  no.~8, pp. 1667--1676, Aug. 2001.

\bibitem{Jensen2004Perceptual}
J.~Jensen, R.~Heusdens, and S.~H. Jensen, ``A perceptual subspace approach for
  modeling of speech and audio signals with damped sinusoids,'' \emph{IEEE
  Trans. Speech Audio Process.}, vol.~12, no.~2, pp. 121--132, Mar. 2004.

\bibitem{Hermus2005Perceptual}
K.~Hermus, W.~Verhelst, P.~Lemmerling, P.~Lemmerling, and S.~V. Huffel,
  ``Perceptual audio modeling with exponentially damped sinusoids,''
  \emph{Signal Process.}, vol.~85, no.~1, pp. 163--176, Jan. 2005.

\bibitem{Patel2016Design}
V.~Patel, V.~Gandhi, S.~Heda, and N.~V. George, ``Design of adaptive
  exponential functional link network-based nonlinear filters,'' \emph{IEEE
  Trans. Circuits Syst. I, Reg. Papers}, vol.~63, no.~9, pp. 1434--1442, Sep.
  2016.

\bibitem{Patel2020Convergence}
V.~Patel, S.~S. Bhattacharjee, and N.~V. George, ``Convergence analysis of
  adaptive exponential functional link network,'' \emph{IEEE Trans. Neural
  Netw. Learn. Syst.}, 2020, in press, DOI: 10.1109/TNNLS.2020.2979688.

\bibitem{Vasundhara2018Decorrelated}
Vasundhara, N.~B. Puhan, and G.~Panda, ``De-correlated improved adaptive
  exponential {FLAF}-based nonlinear adaptive feedback cancellation for hearing
  aids,'' \emph{IEEE Trans. Circuits Syst. I, Reg. Papers}, vol.~65, no.~2, pp.
  650--662, Feb. 2018.

\bibitem{Le2018Generalized}
D.~C. Le, J.~Zhang, D.~Li, and S.~Zhang, ``A generalized exponential functional
  link artificial neural networks filter with channel-reduced diagonal
  structure for nonlinear active noise control,'' \emph{Appl. Acoust.}, vol.
  139, pp. 174--181, Oct. 2018.

\bibitem{Deb2020Design}
T.~Deb, D.~Ray, and N.~V. George, ``Design of nonlinear filters using affine
  projection algorithm based exact and approximate adaptive exponential
  functional link networks,'' \emph{IEEE Trans. Circuits Syst. II, Exp.
  Briefs}, vol.~67, no.~11, pp. 2757--2761, Nov. 2020.

\bibitem{Bhattacharjee2020Nonlinear}
S.~S. Bhattacharjee and N.~V. George, ``Nonlinear system identification using
  exact and approximate improved adaptive exponential functional link
  networks,'' \emph{IEEE Trans. Circuits Syst. II, Exp. Briefs}, vol.~67,
  no.~12, pp. 3542--3546, Dec. 2020.

\bibitem{Yu2020DCD}
Y.~Yu, L.~Lu, Z.~Zheng, W.~Wang, Y.~Zakharov, and R.~C. de~Lamare,
  ``{DCD}-based recursive adaptive algorithms robust against impulsive noise,''
  \emph{IEEE Trans. Circuits Syst. II, Exp. Briefs}, vol.~67, no.~7, pp.
  1359--1363, Jul. 2020.

\bibitem{Yu2021Robust}
T.~Yu, W.~Li, Y.~Yu, and R.~C. de~Lamare, ``Robust spline adaptive filtering
  based on accelerated gradient learning: {Design} and performance analysis,''
  \emph{Signal Process.}, vol. 183, Jun. 2021, {Art.} no. 107965.

\bibitem{Zheng2020Steady}
Z.~Zheng and Z.~Liu, ``Steady-state mean-square performance analysis of the
  affine projection sign algorithm,'' \emph{IEEE Trans. Circuits Syst. II, Exp.
  Briefs}, vol.~67, no.~10, pp. 2244--2248, Oct. 2020.

\bibitem{Zayyani2014Continuous}
H.~Zayyani, ``Continuous mixed $p$-norm adaptive algorithm for system
  identification,'' \emph{IEEE Signal Process. Lett.}, vol.~21, no.~9, pp.
  1108--1110, Sep. 2014.

\bibitem{Zhou2011New}
Y.~Zhou, S.~C. Chan, and K.~L. Ho, ``New sequential partial-update least mean
  {M}-estimate algorithms for robust adaptive system identification in
  impulsive noise,'' \emph{IEEE Trans. Ind. Electron.}, vol.~58, no.~9, pp.
  4455--4470, Sep. 2011.

\bibitem{Wang2020Constrained}
Z.~Wang, H.~Zhao, and X.~Zeng, ``Constrained least mean {M}-estimation adaptive
  filtering algorithm,'' \emph{IEEE Trans. Circuits Syst. II, Exp. Briefs},
  2020, in press, DOI: 10.1109/TCSII.2020.3022081.

\bibitem{Song2013Normalized}
I.~Song, P.~Park, and R.~W. Newcomb, ``A normalized least mean squares
  algorithm with a step-size scaler against impulsive measurement noise,''
  \emph{IEEE Trans. Circuits Syst. II, Exp. Briefs}, vol.~60, no.~7, pp.
  442--445, Jul. 2013.

\bibitem{Chen2014Steady}
B.~Chen, L.~Xing, J.~Liang, N.~Zheng, and J.~C. Pr\'{\i}ncipe, ``Steady-state
  mean-square error analysis for adaptive filtering under the maximum
  correntropy criterion,'' \emph{IEEE Signal Process. Lett.}, vol.~21, no.~7,
  pp. 880--884, Jul. 2014.

\bibitem{Zhang2018Recursive}
S.~Zhang and W.~X. Zheng, ``Recursive adaptive sparse exponential functional
  link neural network for nonlinear {AEC} in impulsive noise environment,''
  \emph{IEEE Trans. Neural Netw. Learn. Syst.}, vol.~29, no.~9, pp. 4314--4323,
  Sep. 2018.

\bibitem{Petersen2020Topological}
P.~Petersen, M.~Raslan, and F.~Voigtlaender, ``Topological properties of the
  set of functions generated by neural networks of fixed size,'' \emph{Found.
  Comput. Math.}, 2020, in press, DOI: 10.1007/s10208-020-09461-0.

\end{thebibliography}

\end{document}